# Leveraging BERT for Extractive Text Summarization on Lectures

**Derek Miller**
Georgia Institute of Technology
Atlanta, Georgia
dmiller303@gatech.edu

**ABSTRACT**
In the last two decades, automatic extractive text summarization on lectures has demonstrated to be a useful tool for collecting key phrases and sentences that best represent the content. However, many current approaches utilize dated approaches, producing sub-par outputs or requiring several hours of manual tuning to produce meaningful results. Recently, new machine learning architectures have provided mechanisms for extractive summarization through the clustering of output embeddings from deep learning models. This paper reports on the project called "lecture summarization service", a python-based RESTful service that utilizes the BERT model for text embeddings and K-Means clustering to identify sentences closest to the centroid for summary selection. The purpose of the service was to provide student's a utility that could summarize lecture content, based on their desired number of sentences. On top of summary work, the service also includes lecture and summary management, storing content on the cloud which can be used for collaboration. While the results of utilizing BERT for extractive text summarization were promising, there were still areas where the model struggled, providing future research opportunities for further improvement. All code and results can be found here: https://github.com/dmmiller612/lecture-summarizer.

**ACM Classification Keywords**
I.2.7. Natural Language Processing

**INTRODUCTION**
When approaching automatic text summarization, there are two different types: abstractive and extractive. In the case of abstractive text summarization, it more closely emulates human summarization in that it uses a vocabulary beyond the specified text, abstracts key points, and is generally smaller in size (Genest & Lapalme, 2011). While this approach is highly desirable and has been the subject of many research papers, since it emulates how humans summarize material, it is difficult to automatically produce, either requiring several GPUs to train over many days for deep learning or complex algorithms and rules with limited generalizability for traditional NLP approaches. With this challenge in mind, the lecture summarization service uses extractive summarization. In general, extractive text summarization utilizes the raw structures, sentences, or phrases of the text and outputs a summarization, leveraging only the content from the source material. For the initial implementation of the service, only sentences are used for summarization.

In education, automatic extractive text summarization of lectures is a powerful tool, extrapolating the key points without manual intervention or labor. In the context of many MOOCs, transcripts from video lectures are available, but the most valuable information from each lecture can be challenging to locate. Currently, there have been several attempts to solve this problem, but nearly all solutions implemented outdated natural language processing algorithms, requiring frequent maintenance due to poor generalization. Due to these limitations, many of the summary outputs from the mentioned tools can appear random in its construction of content. In the last year, many new deep learning approaches have emerged, proving state of the art results on many tasks, such as automatic extractive text summarization. Due to the need for more current tools in lecture summarization, the lecture summarization service provides a RESTful API and command line interface (CLI) tool that serves extractive summaries for any lecture transcripts with the goal of proving that the implementation can be expanded to other domains. The following sections will explore the background and related work around lecture summarization, the methodologies used in building the service, the results and metrics of the model, and example summarizations, showing how they compare to commonly used tools such as TextRank.

**BACKGROUND AND RELATED WORK**
In order to provide necessary context to the proposed solution of automatic lecture summarization, it is worth investigating previous research, identifying the pros and cons of each approach. In the early days of lecture searching, many multimedia applications created manual summarizations for each lecture. One example of this is from M.I.T's lecture processing project, where they uploaded a large amount of lectures, including transcripts for keyword searching and a summary of the content in the lecture (Glass, Hazen, Cyphers, Malioutov, Huynh, & Barzilay, 2007). For a limited amount of content, this approach can suffice, but as the data scales, the manual summary process can be inefficient. One motivation for manual summarization in the mid 2000's was due to the poor quality from extractive summary tools. In 2005, researchers created a tool that would automatically extract corporate meeting summaries using simple probabilistic models, but quickly found that the output was far inferior to human constructed summarizations (Murray, Renals, & Carletta, 2005). Due to poor performance with this methodology, it led to several research papers that aimed to improve the process.

**Summarization Improvements**
Lacking the widespread use of deep learning algorithms in 2007, researchers attempted to include rhetorical information into their lecture summaries to help improve summarization performance (Zhang, Chan, & Fung, 2007). While this led to a decent performance gain, it still created sub-par outputs, concluding that the technology had potential but needed further research (Zhang, Chan, & Fung, 2007). Six years later, engineers created an industry product called "OpenEssayist" which would output the topics and key points in a student's essay, aiding the student while they were completing their assignment (Van Labeke, Whitelock, Field, Pulman, & Richardson, 2013). In the product, there were multiple types of summarization options that utilized algorithms such as TextRank for key sentence and keyword extraction (Van Labeke, Whitelock, Field, Pulman, & Richardson, 2013). This demonstrated the usefulness of automatic summarization in the education field, providing helpful topics, sentences, and more from an essay to the student, which differentiated itself from prior research. While a great initial approach, algorithms such as TextRank contain a myopic view of spoken context. Researchers Balasubramanian, Doraisamy, and Kanakarajan built a similar application, leveraging the Naive Bayes algorithm that would determine which phrases and elements of the lectures or slides would be the most descriptive in the summarization of a lecture (Balasubramanian, Doraisamy, & Kanakarajan, 2016). This approach differentiated itself from the previous applications in that it used classification instead of unsupervised learning to create the summaries. Although Naive Bayes has shown some success in the NLP domain, its independent assumption of features can eliminate the broader context of a lecture, potentially creating summaries that lack key components.

While lacking the number of citations as the projects mentioned above, in the last couple of years, there have been a variety of new papers that have attempted to tackle the summarization problem for lectures. In a small section of the book "In Recent Developments in Intelligent Computing, Communication and Devices", the author implemented a video subtitle extraction program that would summarize the multimedia input, utilizing TF-IDF (Garg, 2017). While such approaches may have a decent output, for similar reasons as the Naive Bayes algorithm, TF-IDF struggles in representing complex phrasing, potentially missing key points in a lecture. In 2018, another lecture transcript summarization project was created specifically for MOOCs, which had a similar objective as the lecture summarization service, creating a basic probabilistic algorithm that achieved a precision of 60 percent when comparing to manual summarizations (Che, Yang, & Meinel, 2018). While not the best performance from the previously mentioned algorithms, it was the first project that specifically focused on MOOCs, supplying some prior history to the domain.

In more recent literature, there have been several attempts at lecture summarization without lecture transcripts. Two popular techniques have been extracting text from whiteboards or slide decks, then utilizing that information to create a summary. In one research project, the authors created a tool that utilized deep learning to extract written content from the whiteboard and convert it to a text format for further summarization (Kota, Davila, Stone, Setlur, & Govindaraju, 2018). While no deep learning was performed on the lecture transcripts themselves, this was one of the first found research projects that utilized some sort of deep learning algorithm to extract information for lecture summarization. In a project focused around extracting information from slides, the authors utilized both video and audio processing tools to retrieve content, then implemented a TF-IDF to extract keywords and phrases for the final summarization (Shimada, Okubo, Yin, & Ogata, 2018). As mentioned with Kota et al.'s research, the authors used more state-of-the-art approaches for the initial extraction but ended up selecting traditional NLP algorithms for final summarization.

**Moving Towards Deep Learning**
While highlighting all of the above research projects did not implement deep learning for the lecture summarization on transcripts, even for the more modern projects, there were plethora amount of reasons to not use it. Until recently, the recurrent neural network (using long short term memory networks) was the default approach for many natural language processing applications, requiring massive amounts of data, expensive compute resources, and several hours of training to achieve acceptable results, while suffering from poor performance with very long sequences and was prone to overfit (Vaswani, et al., 2017). With this fact in mind, researcher Vaswani presented a superior architecture called the "Transformer", which completely moved away from RNNs and Convolutional Neural Networks (CNN), in favor using an architecture comprised of feed forward networks and attention mechanisms (Vaswani, et al., 2017). While the Transformer architecture alleviated some of the problems with RNNs and CNNs, it still had sub-human performance on many NLP tasks. At the end of 2018, researchers from Google built an unsupervised learning architecture on top of the Transformer architecture called BERT ( Bidirectional Encoder Representations from Transformers) that exceeded nearly all existing models in the NLP space for a wide range of tasks (Devlin, Chang, Lee, & Toutanova, 2018). On top of publishing the results of the model, the research team also published several pre-trained models which could be used for transfer learning on a multitude of different domains and tasks (Devlin, Chang, Lee, & Toutanova, 2018).

Another component missing from previous research project was the feature of dynamic or configurable summary sizes. Users of lecture summarization applications may want to configure the amount of sentences for each lecture summary, providing more or less information based on their needs. Since the BERT model outputs sentence embeddings, these sentences can be clustered with a size of K, allowing dynamic summaries of the lecture (Celikyilmaz, & Hakkani-Tür, 2011). With that in mind, the lecture summarization service implemented the exact same approach, creating dynamic summarizations from taking the centroid sentence in a cluster, rather than static summaries with a fixed size.

## MOTIVATION

Using the background and related work, a missing element to existing research and projects was a lecture summarization service that could be utilized by students with configurable lecture sizes, leveraging the most up to date deep learning research. This fact provided the motivation for the development of the lecture summarization service, a cloud-based service that ran inference from a BERT model to be used for dynamically sized lecture summarizations.

## METHOD

The lecture summarization service comprises of two main components. One feature is the management of lecture transcripts and summarizations, allowing users to create, edit, delete, and retrieve stored items. The other component is the inference from the BERT model to produce embeddings for clustering, using a K-Means model, creating a summary. Below explores each component in detail, outlining the motivation and implementation of the associated features.

### Extractive Text Summarization with BERT and K-Means

When creating summaries from saved lectures, the lecture summarization service engine leveraged a pipeline which tokenized the incoming paragraph text into clean sentences, passed the tokenized sentences to the BERT model for inference to output embeddings, and then clustered the embeddings with K-Means, selecting the embedded sentences that were closest to the centroid as the candidate summary sentences.

### Textual Tokenization

Due to the variability of the quality of text from lecture transcripts, a combination of multiple tokenization techniques was utilized before passing the input to the models. For transcripts derived from Udacity, a custom parser was created to convert data from the ".srt" file format, a special format that contains time stamps for associated phrases, to a standard paragraph form. Once converted, the NLTK library for python was used to extract sentences from the lecture, breaking up the content to be passed into the subsequent models for inference. The final step of text tokenization consisted of removing or editing candidate sentences with the goal of only having sentences that did not need additional context in the final summary. One example of such behavior was removing sentences that had conjunctions at the beginning. On top of these types of sentences, too small or large of sentences were also removed. Another example was removing sentences that mentioned Udacity quizzes. While the removed sentences were rarely selected for the extractive summarization when they were kept in the lecture, they would change the cluster outputs, affecting the centroids which lead to poorer summary candidates. Once these tokenization steps were completed, the content was ready for inference.

### BERT for Text Embedding

Due to its superior performance to other NLP algorithms on sentence embedding, the BERT architecture was selected. BERT builds on top of the transformer architecture, but its objectives are specific for pre-training. On one step, it randomly masks out 10% to 15% of the words in the training data, attempting to predict the masked words, and the other step takes in an input sentence and a candidate sentence, predicting whether the candidate sentence properly follows the input sentence (Devlin, Chang, Lee, & Toutanova, 2018). This process can take several days to train, even with a substantial amount of GPUs. Due to this fact, Google released two BERT models for public consumption, where one had 110 million parameters and the other contained 340 million parameters (Devlin, Chang, Lee, & Toutanova, 2018). Due to the superior performance in the larger pre-trained BERT model, it was ultimately selected for the lecture summarization service.

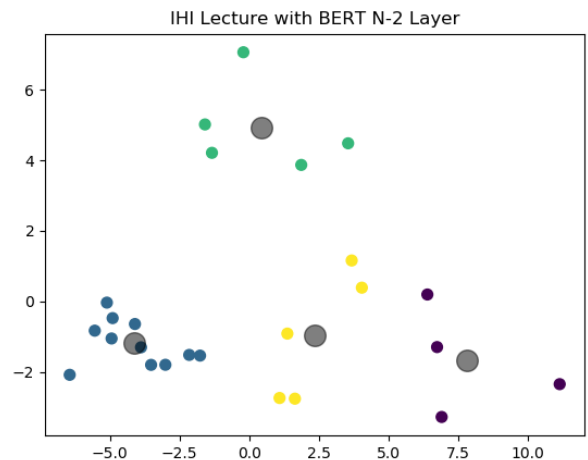

**Figure 1 Introduction to Health Informatics lecture with BERT N-2 layer embeddings**

Using the default pre-trained BERT model, one can select multiple layers for embeddings. Using the [cls] layer of BERT produces the necessary N x E matrix for clustering, where N is the number of sentences and E is the embeddings dimension, but the output of the [cls] layer does not necessarily produce the best embedding representation for sentences. Due to the nature of the BERT architecture, outputs for other layers in the network produced N x W x E embeddings where W equaled the tokenized words. To get around this issue, the embeddings can be averaged or maxed to produce an N x E matrix. After experiments with Udacity extractive summarizations on Udacity lectures, it was determined that the second to last averaged layer produced the best embeddings for representations of words. This was ultimately determined through visual examination of clusters of the initial embedding process. An example of the differences between the two different plots can be seen in both figure 1 and figure 2. Using a sample Introduction to Health Informatics course lecture, one initial hypothesis for the reason of a better sentence representation in the N-2 layer than the final [cls] layer of the BERT network was that the final layer was biased by the classification tasks in the original training of the model.

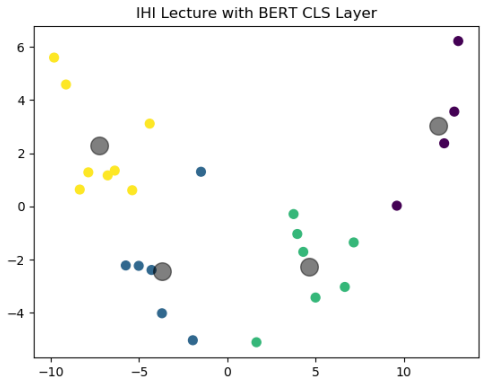

**Figure 2 Introduction to Health Informatics lecture BERT [cls] layer embeddings.**

For the lecture summarization service, the core BERT implementation uses the pytorch-pretrained-BERT library from the "huggingface" organization. At its core, the library is a Pytorch wrapper around Google's pre-trained implementations of the models. On top of the original BERT model, the pytorch-pretrained-BERT library also contains the OpenAi GPT-2 model, which is a network that expands on the original BERT architecture. When examining the sentence embeddings from both the GPT-2 and original BERT model, it was clear that the BERT embeddings were more representative of the sentences, creating larger

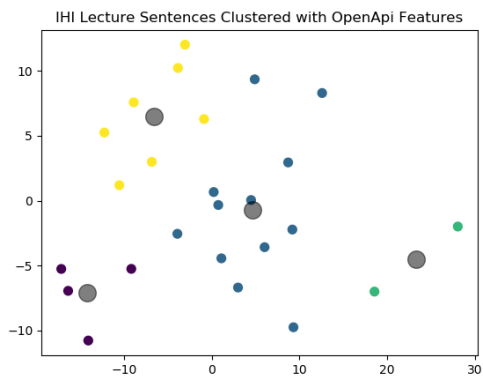

**Figure 3 IHI GPT-2 embeddings**

Euclidean distances between clusters. Below is an example of clustering with the GPT2 embeddings.

**Ensembling Models**
While the OpenAi GPT-2 and BERT embeddings from the [cls] layer provided inferior results, the ensembling of the multiple architectures produced the best results. However, while the clusters had further Euclidean distances from other clusters using this method, its inference time was increased, even when running in a multithreaded environment, requiring a substantial amount of memory and compute as well. With this fact in mind, ensembling was not used in the service as there needed to be a trade-off between inference performance and speed.

**Clustering Embeddings**
Finally, once the N-2 layer embeddings were completed, the N x E matrix was ready for clustering. From the user's perspective, they could supply a parameter K, which would represent the number of clusters and requested sentences for the final summary output. During experimentation, both K-Means and Gaussian Mixture Models were used for clustering, utilizing the Sci-kit Learn library's implementation. Due to models' very similar performance, K-Means was finally selected for clustering incoming embeddings from the BERT model. From the clusters, the sentences closest to the centroids were selected for the final summary.

**Lecture Summarization Service RESTful API**
To provide a sufficient interface to the BERT clustered summarizations, a RESTful API was put in place to serve the models for inference. Since all of the necessary machine learning libraries required python, the Flask library was selected for the server. On top of summarization capabilities, the service also contained lecture transcript management, allowing users to add, edit, delete, and update lectures. This also contained an endpoint which would convert ".srt" files to paragraph text form. Once a lecture was saved into the system, it could be used to run extractive summarizations. Users could supply parameters such as the ratio of sentences to use and a name for a summary to properly organize their resources. Once the summary was completed, they would then be stored in the SQLite database, requiring less compute resources when other users wanted to obtain a summarization of the given lecture. All of the server components were containerized using Docker, so that individuals could run the service locally or deploy it to a cloud provider. Currently, a free-to-use public service exists on AWS, and can be accessed with the following link: http://54.85.20.109:5000. The primary motivation for the RESTful service was to make it extensible for other developers, providing the opportunity for future web applications and command line interfaces to be built on the service.

**Command Line Interface**
While users can directly use the RESTful API for querying the service, a command line interface tool was included for easier interaction. This allows users the ability to upload lecture files from their machine and add it to the service with minimal parameters. Users can also create summaries and list managed resources. The tool can be installed through pip, using the base Github Repository.

## RESULTS

In this section, the focus is on the results of the BERT model and comparing the output to other methodologies such as TextRank. Since there were no golden truth summaries for the lectures, there were no other metrics used besides human comparison and quality of clusters which were discussed in detail in the above sections. Some of the initial weaknesses found in the BERT lecture summarization were the same that other methodologies had, such as sufficiently summarizing large lectures, difficulty in handling context words, and dealing with conversational language over written transcripts, which is more common in lectures.

### Model Weaknesses

For larger lectures, classified as those that have 100 or more sentences, the challenge was to have a small ratio of sentences be properly representative of the entire lecture. When the ratio of sentences to summarize was higher, more of the context was sustained, making it easier to understand the summary for the user. One hypothesis to get around the large lecture issue was to include multiple sentences in a cluster that were closest to the centroid. This would allow more context for the summary, improving the quality of the output. It would also get around the requirement to add more clusters, which could be less representative based on where the centroids converged. The problem with this approach is that it would go directly against the user's ratio parameter, adding more sentences than requested and degrading the user experience with the tool. For this reason, the methodology was not included in the service.

Another weakness with the current approach was that it would occasionally select sentences that contained words that needed further context, such as "this", "those", "these", and "also". While a brute force solution would be to remove sentences that contain these words, quite frequently this change would dramatically reduce the quality of summarizations. Given more time, one potential solution was to use NLTK to find the parts of speech and attempt to replace pronouns and keywords with their proper values. This was initially attempted, but some lectures contained context words that were referenced two to three sentences in the past, making difficult to determine which item was actually the true context.

### Examples

To get a sense of the performance, it is worth looking at some of the summarized content, then comparing the results to a traditional approach like TextRank. Below represents a few example summaries from the Introduction to Health Informatics (https://classroom.udacity.com/courses/ud448) and Reinforcement Learning (https://classroom.udacity.com/courses/ud600) lectures on Udacity.

### Health Information Exchange: Semantic Interoperability

In this lecture, the subject is all around semantic interoperability in health exchanges. Both summaries below contain five out of the total thirty-four sentences for a single sub-section. The BERT summary better captured the context about the creation of the technology around data governance. However, the TextRank model had the benefit of introducing IHIE in the summary which was beneficial to the user for the background. At the same time, TextRank was inferior in selecting sentences that flowed together in its summaries, selecting candidates that had missing context words and more. While the BERT model also contained sentences needing context, the model was able to collect sentences that supplied broader context. Both outputs agreed on the final sentence, which was introducing Dr. Jon Duke for an interview.

### BERT Output

*"The most sophisticated form of HIE creates semantic interoperability, thereby bridging the many ways the same concepts can be expressed by providers and represented in EHRs and other clinical systems. The Regenstrief Institute in Indiana created the expensive and sophisticated technology used by IHIE. This architecture is convenient for data governance, analysis and reporting. In it, all data is stored centrally but in so-called data lockers that remain under the control of the entity that is the source of the data. Next, we will talk to my Georgia Tech colleague, Dr. Jon Duke, who came here from the Regenstrief Institute that created the technology used in the Indiana Health Information Exchange."*

### TextRank Output

*"The premier example of this here in the U.S. is the Indiana Health Information Exchange or IHIE, pronounced, "I-hi". Support for that came from the Regenstrief Foundation, a philanthropic organization that describes its mission as, "To bring to the practice of medicine the most modern scientific advances from engineering, business and the social sciences; and to foster the rapid dissemination into medical practice of the new knowledge created by research." Absent such unfortunately rare funding source, this type of HIE is usually economically impossible to create. In the case of IHIE, as you see here, all the curated data is aggregated and stored centrally. We built our Georgia Tech FHIR server using the OMOP data model. Next, we will talk to my Georgia Tech colleague, Dr. Jon Duke, who came here from the Regenstrief Institute that created the technology used in the Indiana Health Information Exchange."*

### Reinforcement Learning – TD(0)

In the Reinforcement Learning course, the content is structured in a way that is conversational between two authors. This brings about another challenge, which is summarizing content that is part of a conversation. Below is an example of both BERT and TextRank summarizing this content for a TD(0) lecture, reducing the sentence size from 40 to 5. In this example, the strengths of the BERT model can be seen, as it addresses that the build-up to the maximum likelihood is the equivalent to TD(0). It properly selects the definition of TD(0) as well, and strings together a summary which properly abstracts the data. While TextRank contains the word Maximum Likelihood, the sentences are rather random, making it difficult to understand TD(0) from the content given.

**BERT Output**

*"All right, so here is a rule we're going to call the TD (0) rule, which gives it a different name from TD (1). So the thing that's random here, at least the way we've been talking about it is, if we were in some state St-1 and we make a transition, we don't know what state we're going to end up in. So really we're taking an expectation over what we get as the next state of the reward plus the discounted estimated value of that next state. Yeah, so this is exactly what the maximum likelihood estimate is supposed to be. As long as these probabilities for what the next state is going to be match what the data has shown so far as the transition to State."*

**TextRank Output**

*"That the way we're going to compute our value estimate for the state that we just left, when we make a transition at epoch T for trajectory T, big T, is what the previous value was. So what would we expect this outcome to look like on average, right? Yeah, so here's the idea, is that if we repeat this update rule on the finite data that we've got over and over and over again, then we're actually taking an average with respect to how often we've seen each of those transitions. Kind of everything does the right thing in infinite data. And the issue is that if we run our update rule over that data over and over and over again, then we're going to get the effect of having a maximum likelihood model.*

**FUTURE IMPROVEMENTS**

For model future improvements, one strategy could be to fine-tune the model on Udacity lectures, since the current model is the default pre-trained model from Google. The other improvement would be to fill in the gaps for missing context from the summary, and automatically determine the best number of sentences to represent the lecture. This could be potentially done through the sum of squares with clustering. For the service, the database would eventually need to be converted to a more permanent solution over SQLite. Also, having logins where individuals could manage their own summaries would be another beneficial feature.

**CONCLUSION**

Having the capability to properly summarize lectures is a powerful study and memory refreshing tool for university students. Automatic extractive summarization researchers have attempted to solve this problem for the last several years, producing research with decent results. However, most of the approaches leave room for improvement as they utilize dated natural language processing models. Leveraging the most current deep learning NLP model called BERT, there is a steady improvement on dated approaches such as TextRank in the quality of summaries, combining context with the most important sentences. The lecture summarization service utilizes the BERT model to produce summaries for users, based on their specified configuration. While the service for automatic extractive summarization was not perfect, it provided the next step in quality when compared to dated approaches.